\documentclass[runningheads]{llncs}
\usepackage[T1]{fontenc}
\usepackage{amsmath}
 \usepackage[caption=false]{subfig}
\usepackage[export]{adjustbox}
\usepackage{float}
\usepackage[dvipsnames]{xcolor}
\usepackage{xr}
\usepackage{algorithm}
\usepackage{booktabs}
\usepackage{multirow}
\usepackage{array}
\usepackage{algpseudocode}
\usepackage{booktabs}
\usepackage{graphicx,verbatim}
\usepackage{mathabx}
\usepackage{pifont}
\newcommand{\cmark}{\ding{108}}%
\newcommand{\xmark}{\ding{109}}%
\newcommand{\halfmark}{\ding{119}}
\usepackage{hyperref}
\hypersetup{
    colorlinks=true,
    linkcolor=blue,
    citecolor=blue,
}
\newcommand\blfootnote[1]{%
  \begingroup
  \renewcommand\thefootnote{}\footnote{#1}%
  \addtocounter{footnote}{-1}%
  \endgroup
}

\newcommand{\beginsupplement}{%
        \setcounter{table}{0}
        \renewcommand{\thetable}{S\arabic{table}}%
        \setcounter{figure}{0}
        \renewcommand{\thefigure}{S\arabic{figure}}%
     }
\begin{document}
\title{Robust automatic brain vessel segmentation in 3D CTA scans using dynamic 4D-CTA data}
\titlerunning{Robust brain vessel segmentation using dynamic CTA data}
%

\author{Alberto Mario Ceballos Arroyo\textsuperscript{\textdagger,*,}\inst{1} \and
Shrikanth M. Yadav\textsuperscript{\textdagger,}\inst{2,3,5} \and Chu-Hsuan Lin\inst{4,5} \and Jisoo Kim \inst{4,5} \and Geoffrey S. Young \inst{4,5} \and Lei Qin \inst{3,5}  \and  Huaizu Jiang \inst{1}}
\authorrunning{Ceballos Arroyo, A.M.; Yadav, S.M. \textit{et al.}}
\institute{Northeastern University, Boston, USA \and
Washington University in St. Louis, St, Louis, USA \and Dana-Farber Cancer Institute, Boston, USA \and Brigham and Women's Hospital, Boston, USA \and Harvard Medical School, Boston, USA}
\maketitle              
\begin{abstract}
In this study, we develop a novel methodology for annotating the brain vasculature using dynamic 4D-CTA head scans. By using multiple time points from dynamic CTA acquisitions, we subtract bone and soft tissue to enhance the visualization of arteries and veins, reducing the effort required to obtain manual annotations of brain vessels. We then train deep learning models on our ground truth annotations by using the same segmentation for multiple phases from the dynamic 4D-CTA collection, effectively enlarging our dataset by 4 to 5 times and inducing robustness to contrast phases. In total, our dataset comprises 110 training images from 29 patients and 165 test images from 14 patients. In comparison with two similarly-sized datasets for CTA-based brain vessel segmentation, a nnUNet model trained on our dataset can achieve significantly better segmentations across all vascular regions, with an average mDC of 0.846 for arteries and 0.957 for veins in the TopBrain dataset. Furthermore, metrics such as average directed Hausdorff distance (adHD) and topology sensitivity (tSens) reflected similar trends: using our dataset resulted in low error margins (aDHD of 0.304 mm for arteries and 0.078 for veins) and high sensitivity (tSens of 0.877 for arteries and 0.974 for veins), indicating excellent accuracy in capturing vessel morphology. Our code and model weights are available \href{https://github.com/alceballosa/robust-vessel-segmentation}{online}.


\end{abstract}
\section{Introduction}

\blfootnote{\textdagger \hphantom{.}  co-first authors with equal contribution}
\blfootnote{* corresponding author: ceballosarroyo.a@northeastern.edu}

\begin{figure}[t!] 
   \begin{center}
   \includegraphics[width=1\textwidth]{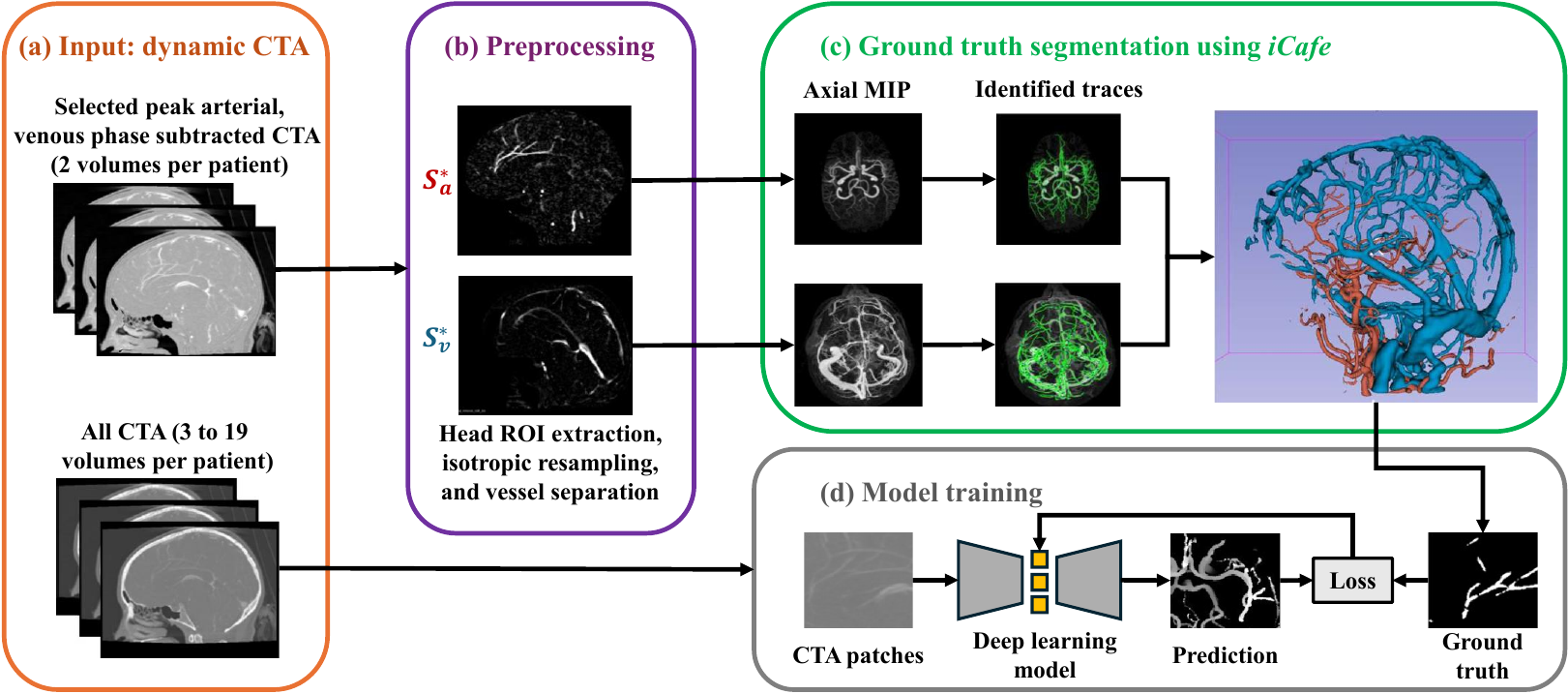}
   
   \caption{Summary of our methodology. (a) Dynamic CTA are collected for all patients. (b) The subtracted images are processed to suppress voxels outside the head and resampled, then the artery-only and vein-only volumes are computed. (c) The vessel-separated images are processed using \textit{iCafe}'s vessel tracing algorithm and then manually validated. (d) For model training, all available CTA data (\textit{i.e.}, 3-19 volumes per patient) are paired with the GT segmentations. 
   \label{fig:methods}     }  
    \end{center}
\end{figure}

Intracranial computed tomographic angiography (CTA) is an essential diagnostic tool that employs intravenous contrast agents to enhance vascular  visualization, aiding in the detection of intracranial vascular abnormalities \cite{walluscheck_mr-ct_2023}. Since its introduction, CTA has advanced beyond single-phase acquisitions to include multiphase CTA \cite{dundamadappa_multiphase_2021}, which captures arterial, mid-venous, and late venous phases, and dynamic 4D-CTA \cite{rajiah_dynamic_2022}, which continuously or intermittently acquires images throughout the passage of contrast to provide detailed temporal vascular information. Despite these advances, single-phase CTA (which requires extensive post-processing to remove background bone/soft tissue and to distinguish arteries from veins) remains the predominant clinical protocol. Robust, automated methods to extract the brain vasculature in single-phase CTA could thus have substantial clinical impact, either by directly assisting radiologists during interpretation of scans or by serving as an auxiliary input to pipelines for tasks such as occlusion localization \cite{kassamDetectingLargeVessel2022,lal-trehanestradaStrategiesCombine3D2024} and aneurysm detection \cite{ceballos-arroyo_vessel-aware_2024,kimAutomatedAnatomybasedHeuristic2025a}.

Most efforts for brain vessel segmentation in CTA have centered on  the automatic segmentation of the Circle of Willis (CoW), an important arterial structure. The TopCoW dataset \cite{yangBenchmarkingCoWTopCoW2025} includes 125 CTA scans which were annotated using a manual, VR-based 3D workflow. The dataset has fine-grained labels for the arteries in the CoW and has drawn significant scientific interest \cite{10.1007/978-3-031-72111-3_68,10.1007/978-3-031-72111-3_5}. Nevertheless, focusing only on the CoW ignores parts of the brain vasculature that are relevant for the management of various cerebrovascular diseases. As a result, two incremental efforts have sprung from TopCoW: a) TopBrain \cite{TopBrainSegmentationChallengea}, which expands the labels in a subset of the scans from TopCoW to include both veins and arteries; and b) VesselVerse \cite{falcettaVesselVerseDatasetCollaborative2026a}, which augments the labels for all of TopCoW with manual and automatic annotations of more arteries and some veins (albeit collapsed into a single vessel class). On the other hand, other researchers have attempted to address the limited availability of annotated CTA data by training deep learning models on large, heterogeneous datasets from multiple modalities. One such example is VesselFM \cite{vesselfm}, which relies on annotations from many modalities (CTA, MRA, X-Ray, microscopy, etc.) combined with a huge number of synthetic images. However, the fraction of training CTA images is small (under 1\%) and the model has not been evaluated for head CTA vessel segmentation. 


\begin{figure}[h] 
   \begin{center}
   \includegraphics[width=1\textwidth]{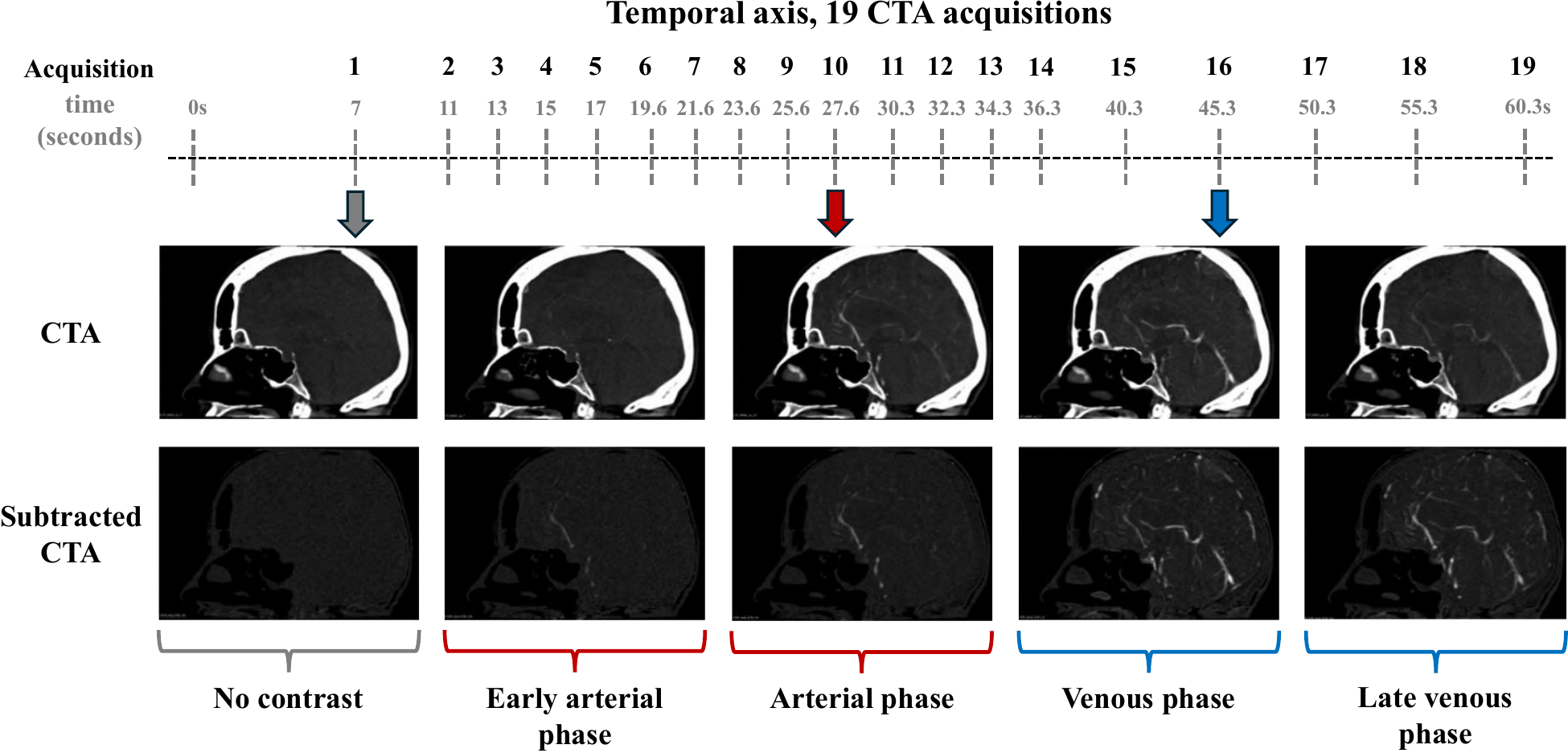}
   
   \caption{Dynamic CTA from a sample patient. The top axis shows the acquisition order of the scans. The red and blue arrows point to the arterial and venous phase images, respectively. The \textit{CTA} row depicts the original CTA images, while the \textit{subtracted CTA} row displays the bone/soft tissue-subtracted versions. 
   \label{fig:data_acquisition}     }  
    \end{center}
\end{figure}

Considering the above, there is a clear need for new annotation methodologies that minimize the time required to produce fine-grained labels for vessels in brain CTA. However, accurate annotation of CTA images requires specialized anatomical knowledge due to the presence of bone, soft tissue, and non-vascular contrast-enhanced structures. To address this challenge, we use dynamic 4D-CTA, which enables near-complete removal of bone and soft tissue through baseline subtraction. The high temporal resolution of dynamic CTA also minimizes interphase motion artifacts, resulting in cleaner subtracted images and more reliable vascular depiction. These enhancements overcome the longstanding difficulties in segmenting and distinguishing intracranial arteries and veins in angiographic CT—a process typically regarded as complex and labor-intensive. Using the proposed strategy (summarized in Fig. \ref{fig:methods}), we annotate a dataset comprising scans from 39 patients with artery and vein labels, and we use it to train an nnUNet \cite{isensee_nnu-net_2021} model, achieving remarkable artery and vein segmentation performance across contrast phases. In summary, our contributions are as follows:

\begin{itemize}
    \item We document a pipeline for annotating arteries and veins in head CTA scans based on dynamic 4D-CTA data, which significantly reduces annotation time
    \item We present DynaVessel, an intracranial vessel segmentation dataset comprising 275 head CTA scans from 43 patients with artery and vein annotations
    \item We evaluate a model trained on DynaVessel against several baselines on various in-domain and out-of-domain benchmarks, achieving substantially better segmentation performance on both arteries and veins
\end{itemize}
\section{Methods}

\subsection{Data}

This retrospective study received IRB approval with a waiver of informed consent.
 The 4D-CTA scans obtained from our clinical research database were acquired using a 320-detector-row CT scanner (Canon-Toshiba Aquilion One, Toshiba Medical Systems, Tokyo, Japan). Imaging was conducted in axial mode using the following parameters: 80 kV, 150 mA, 0.75 s rotation time, 320 slices with 0.5 mm thickness, and a 512 × 512 matrix. For each patient, image acquisition began roughly 7 seconds after injecting 75–100 mL of iodinated contrast agent (Omnipaque 350, General Electric Healthcare, Chicago, USA) at a rate of 4–5 mL/s, continuing intermittently through the baseline, arterial, and venous phases. As shown in Fig.~\ref{fig:data_acquisition}, images were captured at 19 time points over a 60-second interval. The peak arterial and venous phase images, which correspond to $t=27.6$s and $t=45.3$s respectively, are of primary interest. The scanner uses the baseline image—acquired before contrast reaches the brain—to subtract bone and soft tissue from subsequent images. 

\begin{figure}[h] 
   \begin{center}
   \includegraphics[width=0.85\textwidth]{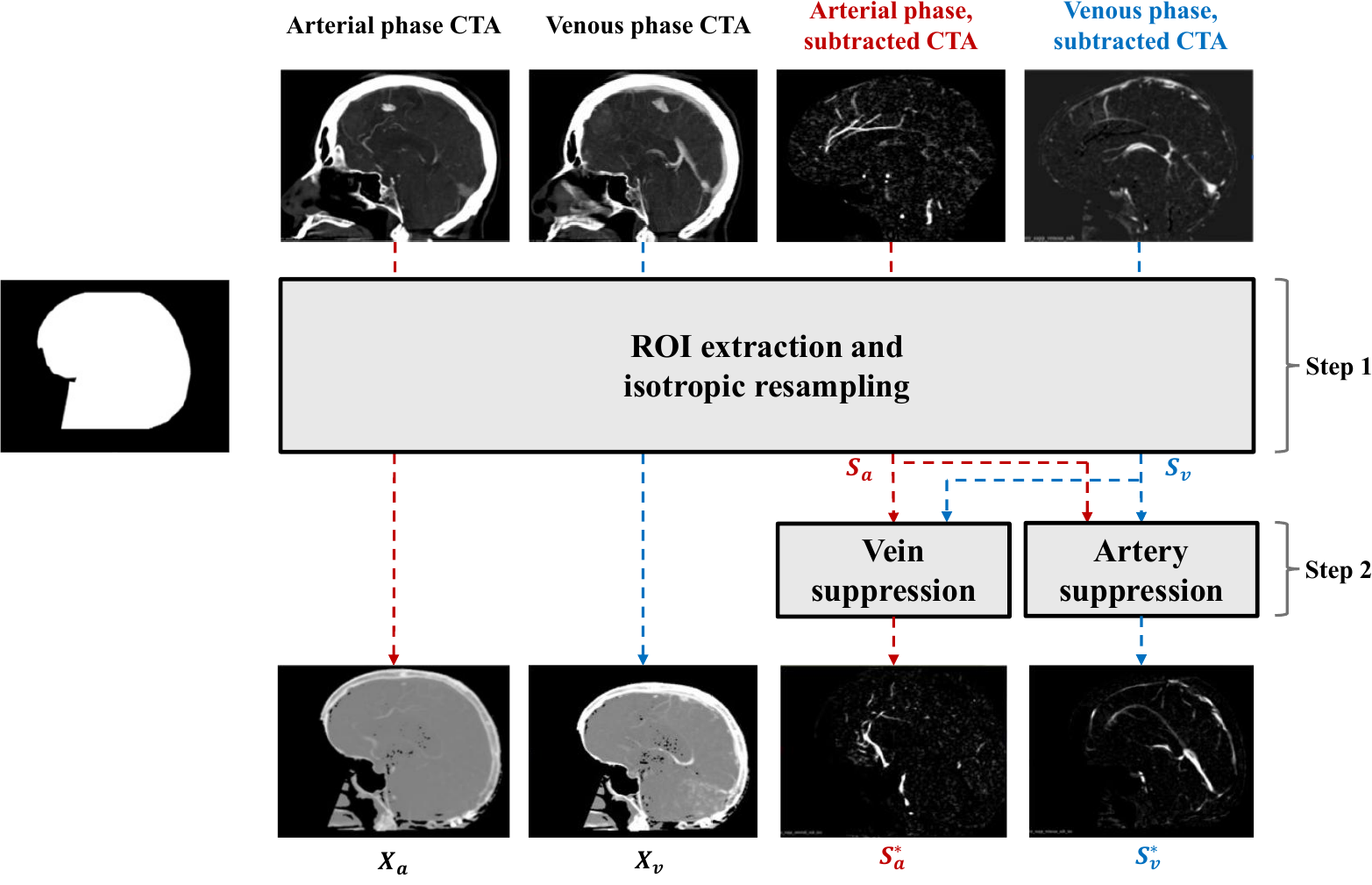}
   \caption{Pre-processing steps for a dynamic CTA study. For both phases, a head ROI is used on the CTA and the subtracted CTA to suppress the background. Next, the subtracted images are processed using Algorithm 1 to create the vessel-separated images. \label{fig:data_preprocessing}     }  
    \end{center}
\end{figure}

Of the 106 patients scanned in 2023, 38 had scans containing at least two phases of acceptable image quality. Although these were acquired using dynamic CTA protocols, complete phase datasets were not consistently archived in the PACS; in many cases, only thicker slice reconstructions, subtracted volumes, or MIP images were retained. After review, 29 patients were identified as having artifact-free arterial and venous phase images. These scans were annotated and used for model training. For evaluation, we retrieved data from 1910 patients scanned between 2016 to 2022, of whom 140 had dynamic CTA with at least two phases of acceptable resolution. Among these, 68 patients had both peak arterial and venous phase images (each comprising the original post-contrast CT image and the corresponding bone/soft tissue-subtracted image), with many also including more CTA phases. A subset of 14 patients were randomly selected to evaluate the performance of deep learning (DL) models trained using our dataset.

\begin{table}[]
\centering
\scriptsize
\caption{Train-test splits for the datasets used in this paper. Coverage for contrast phases (Contr.), proximal arteries (Pr.), first (1O) and second (2O) order arteries, distal arteries (Di.), arterial branches (Br.) and veins (V.) is indicated as \cmark:  complete or near-complete, \halfmark: partial, or \xmark: absent. \label{tab:splits}} 
\begin{tabular}{lcccccccccc}
\hline
\multirow{2}{*}{\textbf{Dataset}} & \multirow{2}{*}{\textbf{Num. train}} & \multirow{2}{*}{\textbf{Num. test}} & \multirow{2}{*}{\textbf{Labels}} & \multicolumn{7}{c}{\textbf{Coverage}}                                                                                                                                                                                                                                      \\ \cline{5-11} 
                                  &                                              &                                        &                                  & \multicolumn{1}{c}{\textbf{Contr.}} & \multicolumn{1}{c}{\textbf{Pr.}} & \multicolumn{1}{c}{\textbf{1O}} & \multicolumn{1}{c}{\textbf{2O}} & \multicolumn{1}{c}{\textbf{Di.}} & \multicolumn{1}{c}{\textbf{Br.}} & \textbf{V.}\\ \hline
VesselFM~\cite{vesselfm}                          & 510,902                                      & --                                     & Vessel-only                      & \multicolumn{1}{c}{--}                      & \multicolumn{1}{c}{--}             & \multicolumn{1}{c}{--}                   & \multicolumn{1}{c}{--}                   & \multicolumn{1}{c}{--}              & \multicolumn{1}{c}{--}                & --             \\ \hline
TopCoW~\cite{yangBenchmarkingCoWTopCoW2025}                             & 112                                          & 13                                     & 13 arteries                      & \multicolumn{1}{c}{\xmark}                    & \multicolumn{1}{c}{\cmark}         & \multicolumn{1}{c}{\cmark}               & \multicolumn{1}{c}{\xmark}                 & \multicolumn{1}{c}{\xmark}            & \multicolumn{1}{c}{\xmark}              & \xmark           \\ \hline
ISLES~\cite{riedel2025isles24realworldlongitudinal,yangBenchmarkingCoWTopCoW2025}                             & --                                           & 26                                     & 13 arteries                      & \multicolumn{1}{c}{\xmark}                    & \multicolumn{1}{c}{\cmark}         & \multicolumn{1}{c}{\cmark}               & \multicolumn{1}{c}{\xmark}                 & \multicolumn{1}{c}{\xmark}            & \multicolumn{1}{c}{\xmark}              & \xmark           \\ \hline
VesselVerse~\cite{falcettaVesselVerseDatasetCollaborative2026a}                       & 112                                          & 13                                     & Vessel-only                      & \multicolumn{1}{c}{\xmark}                    & \multicolumn{1}{c}{\cmark}         & \multicolumn{1}{c}{\cmark}               & \multicolumn{1}{c}{\cmark}               & \multicolumn{1}{c}{\xmark}            & \multicolumn{1}{c}{\xmark}              & \halfmark \\ \hline
TopBrain~\cite{TopBrainSegmentationChallengea,yangBenchmarkingCoWTopCoW2025}                           & --                                           & 25                                     &  6 veins / 34 arteries          & \multicolumn{1}{c}{\xmark}                    & \multicolumn{1}{c}{\cmark}         & \multicolumn{1}{c}{\cmark}               & \multicolumn{1}{c}{\cmark}               & \multicolumn{1}{c}{\cmark}          & \multicolumn{1}{c}{\cmark}            & \cmark         \\ \hline
DynaVessel $_{train}$            & 110                                          & --                                     & Vein / artery                    & \multicolumn{1}{c}{\cmark}                  & \multicolumn{1}{c}{\cmark}         & \multicolumn{1}{c}{\cmark}               & \multicolumn{1}{c}{\cmark}               & \multicolumn{1}{c}{\cmark}          & \multicolumn{1}{c}{\cmark}    & \cmark         \\ \hline
DynaVessel$_1$                     & --                                           & 14                                     & Vein / 22 arteries                    & \multicolumn{1}{c}{\xmark}                    & \multicolumn{1}{c}{\cmark}         & \multicolumn{1}{c}{\cmark}               & \multicolumn{1}{c}{\cmark}               & \multicolumn{1}{c}{\halfmark}            & \multicolumn{1}{c}{\halfmark}              & \cmark         \\ \hline
DynaVessel$_+$                     & --                                           & 165                                    & Vein / 22 arteries            & \multicolumn{1}{c}{\cmark}                  & \multicolumn{1}{c}{\cmark}         & \multicolumn{1}{c}{\cmark}               & \multicolumn{1}{c}{\cmark}               & \multicolumn{1}{c}{\halfmark}            & \multicolumn{1}{c}{\halfmark}              & \cmark         \\ \hline
\end{tabular}
\end{table}

\subsubsection{Pre-processing}



Fig.~\ref{fig:data_preprocessing} illustrates the pre-processing steps for annotation. To suppress extracranial soft tissues, a binary ROI encompassing the skull and neck was delineated on a publicly available CT template \cite{talou_adaptive_2021}. The template was affine registered to the patient’s CTA image, and the resulting transformation matrix was applied to the ROI to create a binary mask of the skull and neck in the patient’s space. Voxels outside of this ROI were excluded from further analysis. All images were then resampled to achieve an isotropic resolution of 0.468 mm. Algorithm 1 was applied to the subtracted CTA images to separate arterial and venous structures into distinct volumes for each patient.  The inputs to the algorithm include the bone/soft tissue-subtracted arterial ($S_a$) and venous ($S_v$) phase images (generated by the scanner), along with the corresponding CTA images $X_a$ and $X_v$. We use affine transformations (steps 1 and 2) to align the images and minimize motion artifacts between acquisitions. Steps 5 and 6 enhance the contrast-to-background ratio for arteries and veins, respectively. The output of Algorithm 1 for each patient is two volumes: $S^*_a$ (bone- and vein-suppressed arterial phase) and $S^*_v$ (bone- and artery-suppressed venous phase).

\begin{algorithm}
\scriptsize
\caption{Artery and vein suppression.} 
   \textbf{Input:} $S_a$, $S_v$, $X_a$, $X_v$ \\
 \textbf{Output:} $S^*_a$, $S^*_v$ \\
     \textbf{1:} Compute rigid transformation $G_{rv}$ such that $X_a$ is moving and $X_v$ is fixed\\
\textbf{2:} Compute rigid transformation $G_{ra}$ such that $X_v$ is moving and $X_a$ is fixed \\
 \textbf{3:} $S_{v \rightarrow a} \gets G_{ra}(X_v)$, \text{transform $X_v$ so that it is in the same space as $X_a$} \\
 \textbf{4:} $S_{a \rightarrow v} \gets G_{rv}(X_a)$, \text{transform $X_a$ so that it is in the same space as $X_v$} \\
     \textbf{5:} $s^*_a \gets \begin{cases} 
    0 & \text{if } s_a < s_{v \rightarrow a} \\
    s_a & \text{if } s_a > s_{v \rightarrow a} 
    \end{cases}$, \text{where $s^*_a$, $s_{v \rightarrow a}$, and $s_a$ are elements of matrices $S^*_a$, $S_{v \rightarrow a}$, and $S_a$}
     \textbf{6:} $s^*_v \gets \begin{cases} 
    0 & \text{if } s_v < s_{a \rightarrow v} \\
    s_v & \text{if } s_v > s_{a \rightarrow v} 
    \end{cases}$, \text{where $s^*_v$, $s_{a \rightarrow v}$, and $s_v$ are elements of matrices $S^*_v$, $S_{a \rightarrow v}$, and $S_v$}
\end{algorithm}

\subsubsection{Ground truth (GT) segmentation}

Based on the vessel-separated volumes generated from the pre-processing steps in Figure~\ref{fig:data_preprocessing}, we label the arteries and veins using the \textit{iCafe} \cite{iCafe}  feature extraction pipeline. This process uses intensity normalization followed by segmentation using the Phansalkar threshold \cite{phansalkar_adaptive_2011} (used for vessel tracking) and the Renyi entropy threshold \cite{kapur_new_1985} (used for visualization). A vessel tracing algorithm,  based on a open-curve active contour model, is used to identify vessel-like structures \cite{wang_broadly_2011}. Spurious vessel traces outside the skull and within the soft tissue of the head are then manually deleted. Finally, the vessel trace and its corresponding mask were exported as a 3D segmentation of the vascular structures. Fig.~\ref{fig:methods}(c) shows the overall process of extracting the GT, which results in a two-class (artery, vein) segmentation dataset.  

\begin{figure}[h] 
   \begin{center}
   \includegraphics[width=1\textwidth]{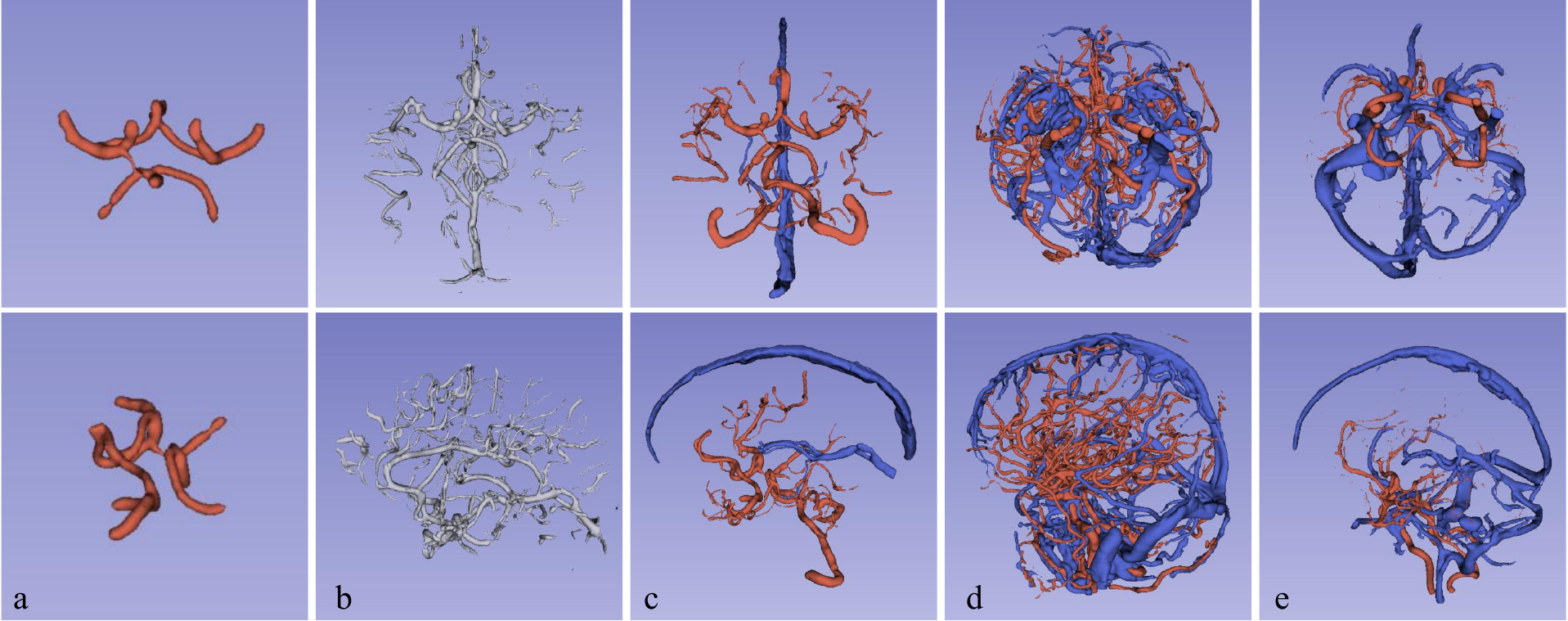}
   \caption{Comparison of vascular structure coverage across datasets, with arteries in red and veins in blue: (a) TopCow, (b) VesselVerse, (c) TopBrain, (d) DynaVessel train, (e) DynaVessel test. Visualizations created on 3D Slicer. \label{fig:coverage}     }  
    \end{center}
\end{figure}

For the evaluation dataset, 20 arteries were annotated using \textit{iCafe}: ICA (Left), ICA (Right), VA (Left), VA (Right), BA, PComm (Left), PComm (Right), P1 (Left), P1 (Right), P2+ (Left), P2+ (Right), M1 (Left), M1 (Right), M2+ (Left), M2+ (Right), A1 (Left), A1 (Right), A2+ (Left), A2+(Right), and Acomm. A subspecialty-trained neuroradiologist validated the annotations by overlaying the segmentations with the subtracted images $S_a$ in  \href{https://www.slicer.org}{3D Slicer} \cite{Kikinis2014} and corrected inaccuracies related to anatomical variations in the arteries. \textit{iCafe} lacks a dedicated venous annotation pipeline, so the GT vessel segmentations of the veins for the evaluation dataset were manually refined by using 3D Slicer to remove distal branches from the venous segmentation. Fig.~\ref{fig:coverage} depicts segmentations from our train (d) and test (e) data, compared with three other datasets (a, b, c).   

\subsection{Deep learning model training} \label{sec:model_training}

Taking advantage of dynamic CTA images, our training and evaluation data includes multiple CTA phases from the same patient, each paired with a corresponding GT segmentation. All images were rigidly registered to the arterial phase CTA using Advanced Normalization Tools~\cite{tustison_antsx_2021}. We split the training and test datasets at the patient level, to prevent any patient from appearing in both validation and training sets. To train the nnUNet model, we use the nnUNet v2 library~\cite{isensee_nnu-net_2021,10.1007/978-3-031-72114-4_47}. Aside from specifying the model type to be a Residual Encoder UNet (large preset), all models were trained for 1000 epochs with a linearly decaying learning rate starting at 0.01, using the nnUNet planner's automatically selected hyper-parameters. All model training and evaluations were performed on a workstation with an NVIDIA RTX 4090 GPU (24 GB VRAM), 8 CPU cores, and 64 GB of RAM.

\subsection{Evaluation} \label{sec:eval}

 Arterial and venous structures are evaluated separately to highlight potential differences in prediction accuracy across vessel types. Importantly, the DL models trained on our dataset output a volume with two classes (artery and vein), but the evaluation dataset contains 21 classes. Although we cannot provide a comprehensive semantic segmentation comparison, we analyze how the predictions align with different vascular structures to provide valuable insight into region-specific performance. Specifically, let volume $A_i$ represent the ground truth vessel mask of interest, where 
$i=1,2\ldots20$ corresponds to arterial labels and $i=21$ corresponds to venous labels; let volume $P$ denote the predicted mask for the arterial or venous region. In line with established guidelines \cite{yangBenchmarkingCoWTopCoW2025}, we use the following counting, distance, and centerline-based metrics: 

\begin{enumerate}
    \item \textbf{Modified Dice coefficient}: we propose a modified version of the Dice coefficient (mDC) focusing on sensitivity, defined as:
    \[
    mDC(A_i,P) = \frac{|A_i \cap P|}{|A_i|}
    \]
    \item \textbf{Average directed Hausdorff surface distance}: this variation of the Hausdorff distance was chosen based on previous work ~\cite{aydin_usage_2021}. $adHD$ is similar to the balanced Hausdorff distance, but we use only the first term in the equation. This is because, in our application, the boundary of the predicted volume can be significantly larger than the GT. Thus, $adHD$ is defined as:
    \[
    adHD(A_{s,i}, P_s) = \frac{1}{|A_{s,i}|} \sum_{a \in A_{s,i}} \min_{p \in P_s} d(a, p)
    \]
    where $A_{s,i}$ and $P_s$ denote the surfaces of volumes $A_i$ and $P$, respectively. $d(a, p)$ represents the Euclidean distance between surface points.
    \item \textbf{Topology sensitivity}: finally, we compute topology sensitivity ($tSens$) to quantify the extent of centerlines that are captured by the vessel segmentation models. This metric is defined based on \textit{clDice}~\cite{shit_cldice_2021}:
    \[
    tSens(A_{c,i}, P) = \frac{|A_{c,i} \cap P|}{|A_{c,i}|}
    \]
    where $A_{c,i}$ is the centerline volume of $A_i$.
\end{enumerate}

\begin{table}[t]
\centering
\scriptsize
\caption{Binary vessel segmentation results on three test sets.} \label{tab:results_1p}
\label{tab:binary_vessel_results}
\begin{tabular}{ll@{\hspace{1em}}l@{\hspace{1.5em}}c@{\hspace{1.5em}}c@{\hspace{1.5em}}c}
\toprule
\textbf{Test set} & \textbf{Model} & \textbf{Train set} & \textbf{mDC} $\uparrow$& \textbf{tSens} $\uparrow$& \textbf{adHD} $\downarrow$ \\
\midrule
\multirow{4}{*}{\textbf{CoW}} & DynUNet & VesselFM & 0.721±0.17 & 0.849±0.14 & 0.566±0.58 \\
 & nnUNet v2 & CoW & 0.789±0.03 & 0.930±0.04 & 0.274±0.13 \\
 & nnUNet v2 & VesselVerse & 0.938±0.04 & 0.989±0.01 & 0.052±0.03 \\
 & nnUNet v2 & DynaVessel & \textbf{0.972}±0.02 & \textbf{0.992}±0.01 & \textbf{0.026}±0.01 \\
\midrule
\multirow{4}{*}{\textbf{ISLES}} & DynUNet & VesselFM & 0.744±0.18 & 0.880±0.16 & 0.572±1.11 \\
 & nnUNet v2 & CoW & 0.723±0.07 & 0.912±0.05 & 0.458±0.28 \\
 & nnUNet v2 & VesselVerse & 0.927±0.03 & 0.987±0.02 & 0.075±0.04 \\
 & nnUNet v2 & DynaVessel & \textbf{0.960}±0.03 & \textbf{0.989}±0.02 & \textbf{0.041}±0.03 \\
\midrule
\multirow{4}{*}{\textbf{VesselVerse}} & DynUNet & VesselFM & 0.320±0.09 & 0.322±0.08 & 3.266±0.94 \\
 & nnUNet v2 & CoW & 0.123±0.04 & 0.079±0.02 & 23.36±2.85 \\
 & nnUNet v2 & VesselVerse & 0.506±0.10 & 0.540±0.10 & 2.757±1.62 \\
 & nnUNet v2 & DynaVessel & \textbf{0.650}±0.08 & \textbf{0.682}±0.09 & \textbf{0.704}±0.35 \\
\bottomrule
\end{tabular}
\end{table}


    \begin{figure*}[h!]
        
        \subfloat{
            \includegraphics[width=.5\linewidth]{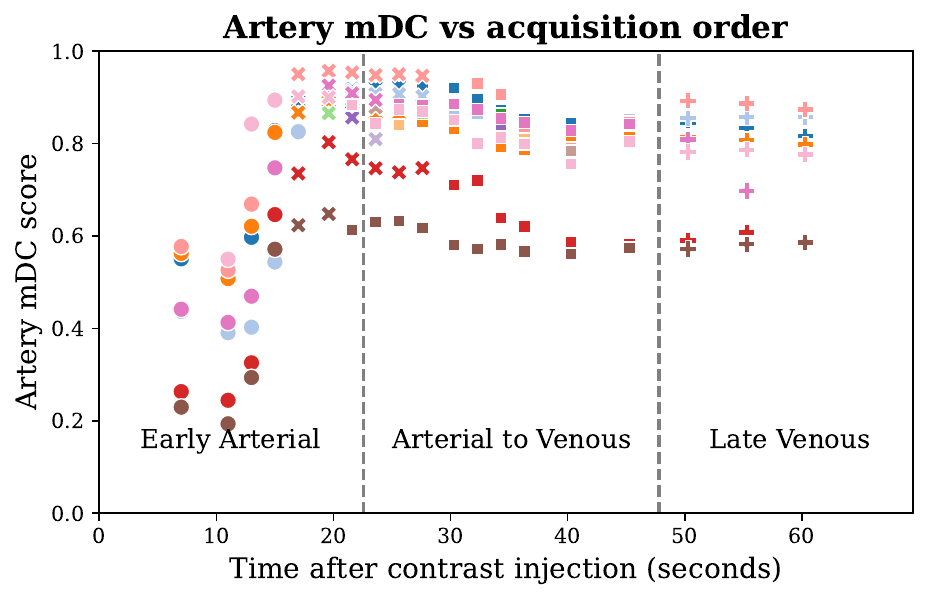}%
            \label{subfig:a}%
        }\hfill
        \subfloat{
            \includegraphics[width=.5\linewidth]{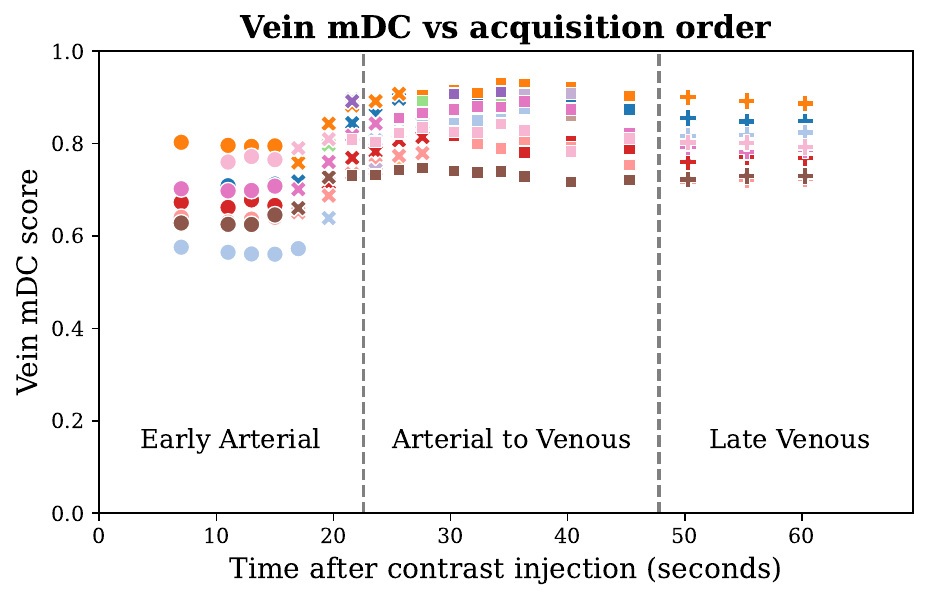}%
            \label{subfig:b}%
        }\\
        \subfloat{
            \includegraphics[width=.5\linewidth]{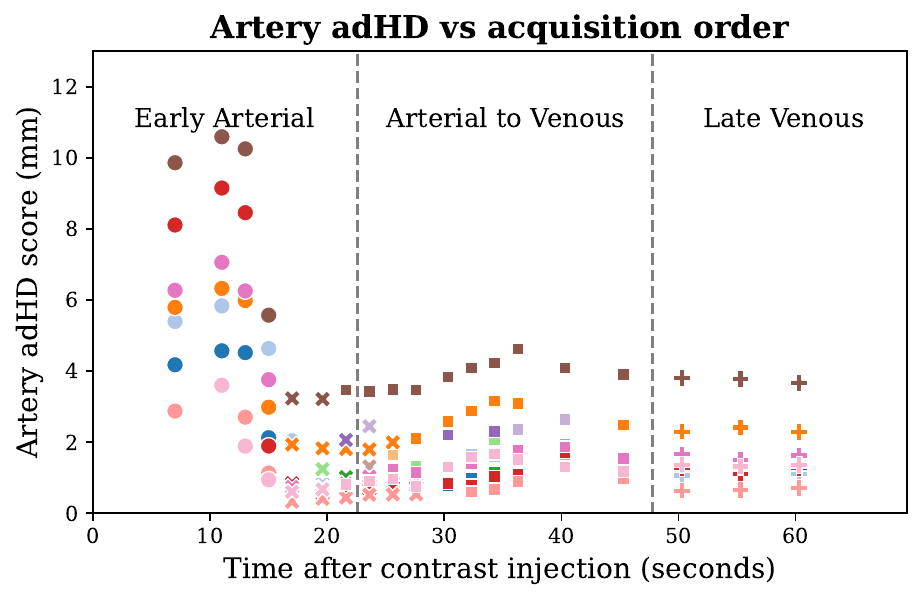}%
            \label{subfig:c}%
        }\hfill
        \subfloat{
            \includegraphics[width=.5\linewidth]{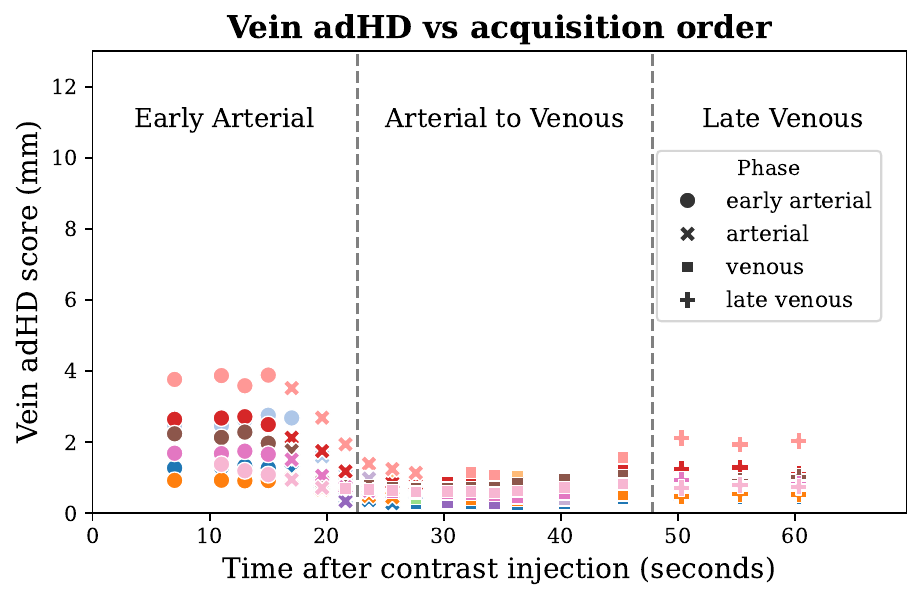}%
            \label{subfig:d}%
        }

        \caption{mDC and adHD scores across phases. Each point represents a single scan from a patient's  Dynamic 4D-CTA, with color encoding the patient. Since the transition between arterial and venous phases is not discrete, we classify an scan's phase as arterial vs venous by taking the average HU intensity on the centerline of the GT artery and vein masks; if the HU is greater for arteries, the scan goes in the arterial phase category, otherwise, it is put in the venous phase category.}
        \label{fig:detailed_phases}
    \end{figure*}
    \begin{figure*}[h!]
        \subfloat{
            \includegraphics[width=1.\linewidth]{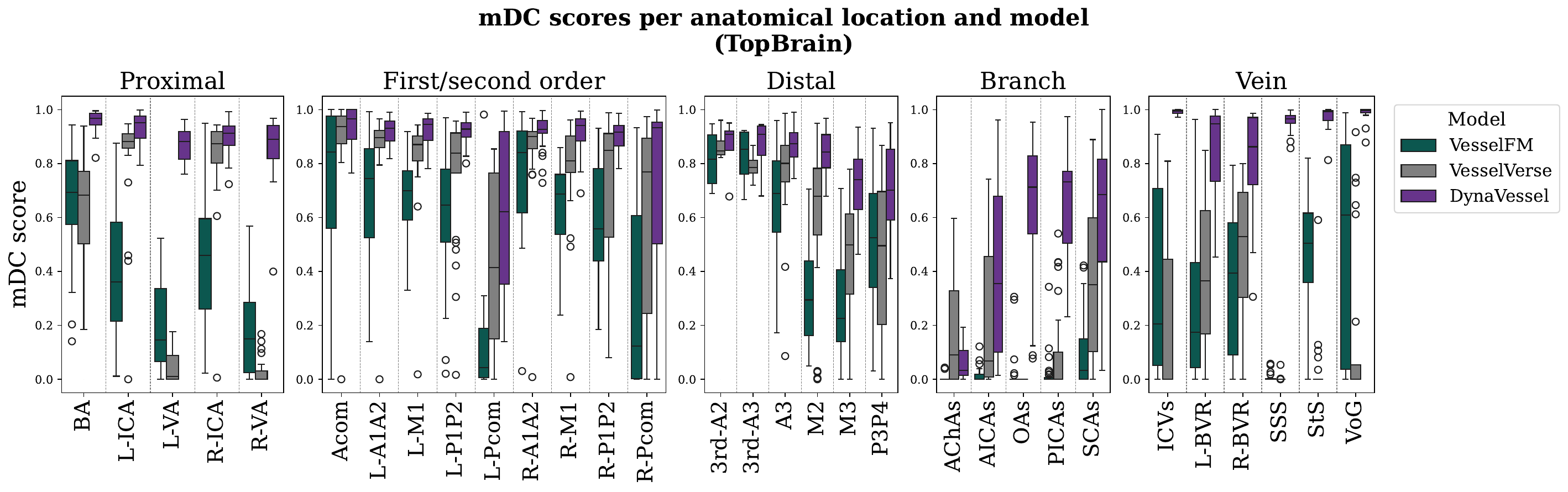}%
            \label{subfig:a}%
        }\\
        \subfloat{
            \includegraphics[width=1\linewidth]{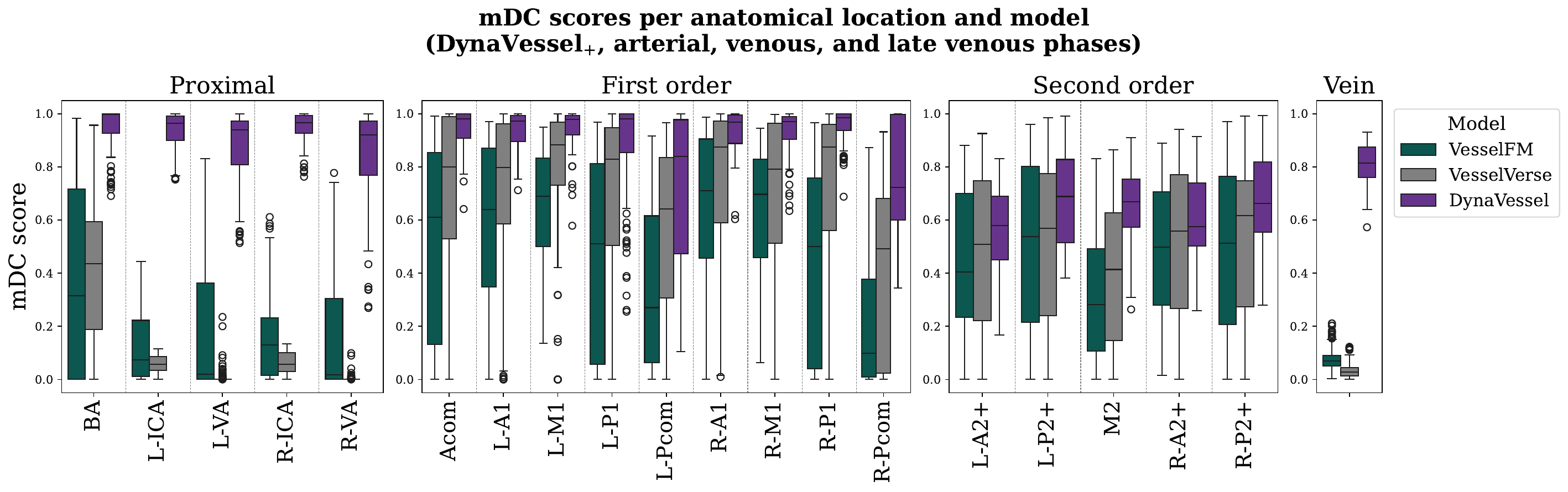}%
            \label{subfig:c}%
        }
        \caption{mDC scores for semantic segmentation. For TopBrain, we merge annotations of right/left distal arteries and branches to improve readability. We omit the pre-contrast/early arterial phase images when calculating mDC on DynaVessel.}
        \label{fig:semantic}
    \end{figure*}

\begin{table}[htbp]
\centering
\scriptsize
\caption{Artery and vein segmentation results. For DynaVessel$_1$ we use a single, arterial phase scan. For DynaVessel$_+$ we evaluate all scans across phases.}
\label{tab:artery_vein_results}
\begin{tabular}{ll@{\hspace{1em}}l@{\hspace{1em}}l@{\hspace{1.5em}}c@{\hspace{1.5em}}c@{\hspace{1.5em}}c}
\toprule
\textbf{Dataset} & \textbf{Region} & \textbf{Model} & \textbf{Train Set} & \textbf{mDC} $\uparrow$ & \textbf{tSens} $\uparrow$ & \textbf{adHD} $\downarrow$\\
\midrule
\multirow{8}{*}{\textbf{TopBrain}} & \multirow{4}{*}{Artery} & DynUNet & VesselFM & 0.374±0.12 & 0.442±0.12 & 2.470±0.94 \\
 & & nnUNet v2 & CoW & 0.192±0.04 & 0.158±0.03 & 21.15±1.92 \\
 & & nnUNet v2 & VesselVerse & 0.455±0.12 & 0.591±0.17 & 8.070±4.29 \\
 & & nnUNet v2 & DynaVessel & \textbf{0.846}±0.05 & \textbf{0.877}±0.06 & \textbf{0.304}±0.15 \\ \cmidrule{2-7}
 & \multirow{4}{*}{Vein} & DynUNet & VesselFM & 0.083±0.07 & 0.210±0.17 & 7.675±2.49 \\
 & & nnUNet v2 & CoW & 0.000±0.00 & 0.000±0.00 & 50.27±8.95 \\
 & & nnUNet v2 & VesselVerse & 0.031±0.04 & 0.132±0.15 & 17.30±7.44 \\
 & & nnUNet v2 & DynaVessel & \textbf{0.957}±0.03 & \textbf{0.974}±0.02 & \textbf{0.078}±0.06 \\
\midrule
\multirow{8}{*}{\textbf{DynaVessel$_{1}$}} & \multirow{4}{*}{Artery} & DynUNet & VesselFM & 0.414±0.09 & 0.494±0.08 & 1.598±0.44 \\
 & & nnUNet v2 & CoW & 0.119±0.03 & 0.078±0.03 & 26.40±2.42 \\
 & & nnUNet v2 & VesselVerse & 0.380±0.05 & 0.519±0.07 & 8.436±2.07 \\
 & & nnUNet v2 & DynaVessel & \textbf{0.862}±0.08 & \textbf{0.615}±0.10 & \textbf{1.353}±0.83 \\
 \cmidrule{2-7}
 & \multirow{4}{*}{Vein} & DynUNet & VesselFM & 0.036±0.02 & 0.082±0.05 & 5.465±0.75 \\ 
 & & nnUNet v2 & CoW & 0.000±0.00 & 0.000±0.00 & 45.00±3.31 \\
 & & nnUNet v2 & VesselVerse & 0.022±0.03 & 0.057±0.06 & 19.29±3.92 \\
 & & nnUNet v2 & DynaVessel & \textbf{0.786}±0.07 & \textbf{0.673}±0.10 & \textbf{0.984}±0.63 \\
\midrule
\multirow{8}{*}{\textbf{DynaVessel$_{+}$}} & \multirow{4}{*}{Artery} & DynUNet & VesselFM & 0.226±0.19 & 0.284±0.22 & 3.751±2.57 \\
 & & nnUNet v2 & CoW & 0.072±0.05 & 0.054±0.04 & 40.42±40.38 \\
 & & nnUNet v2 & VesselVerse & 0.207±0.15 & 0.276±0.21 & 17.71±19.23 \\
 & & nnUNet v2 & DynaVessel & \textbf{0.757}±0.17 & \textbf{0.510}±0.16 & \textbf{2.281}±2.02 \\
 \cmidrule{2-7}
 & \multirow{4}{*}{Vein} & DynUNet & VesselFM & 0.064±0.05 & 0.129±0.09 & 5.137±1.25 \\
 & & nnUNet v2 & CoW & 0.001±0.00 & 0.003±0.01 & 56.49±32.56 \\
 & & nnUNet v2 & VesselVerse & 0.028±0.03 & 0.071±0.06 & 29.12±19.98 \\
 & & nnUNet v2 & DynaVessel & \textbf{0.788}±0.09 & \textbf{0.683}±0.12 & \textbf{1.059}±0.78 \\
\bottomrule
\end{tabular}
\end{table}
\section{Results}

To assess the usefulness of DynaVessel, we compare it to two publicly available CTA vessel segmentation datasets of similar size: TopCoW \cite{yangBenchmarkingCoWTopCoW2025}, which contains 125 scans with semantic annotations of the arteries in the Circle of Willis; and VesselVerse \cite{falcettaVesselVerseDatasetCollaborative2026a}, which expands the annotations in TopCoW to include veins and more arteries, although it uses a single vessel label. We thus train three nnUNet models as specified in Section \ref{sec:model_training} (from now on, we refer to the model trained on DynaVessel as $M_{DV}$). Likewise, we compare against the pre-trained VesselFM \cite{vesselfm} DynUNet, a foundation vessel segmentation model optimized on a collection of 928 images (including TopCoW) from vessel segmentation datasets encompassing various modalities, 10,000 images sampled from a generative flow-matching model, and 500,000 randomly generated 3D images imitating blood vessels based on graph representations of corrosion casts. Similar to VesselVerse, the labels in VesselFM are binary.

\begin{figure}[h] 
   \begin{center}
   \includegraphics[width=1\textwidth]{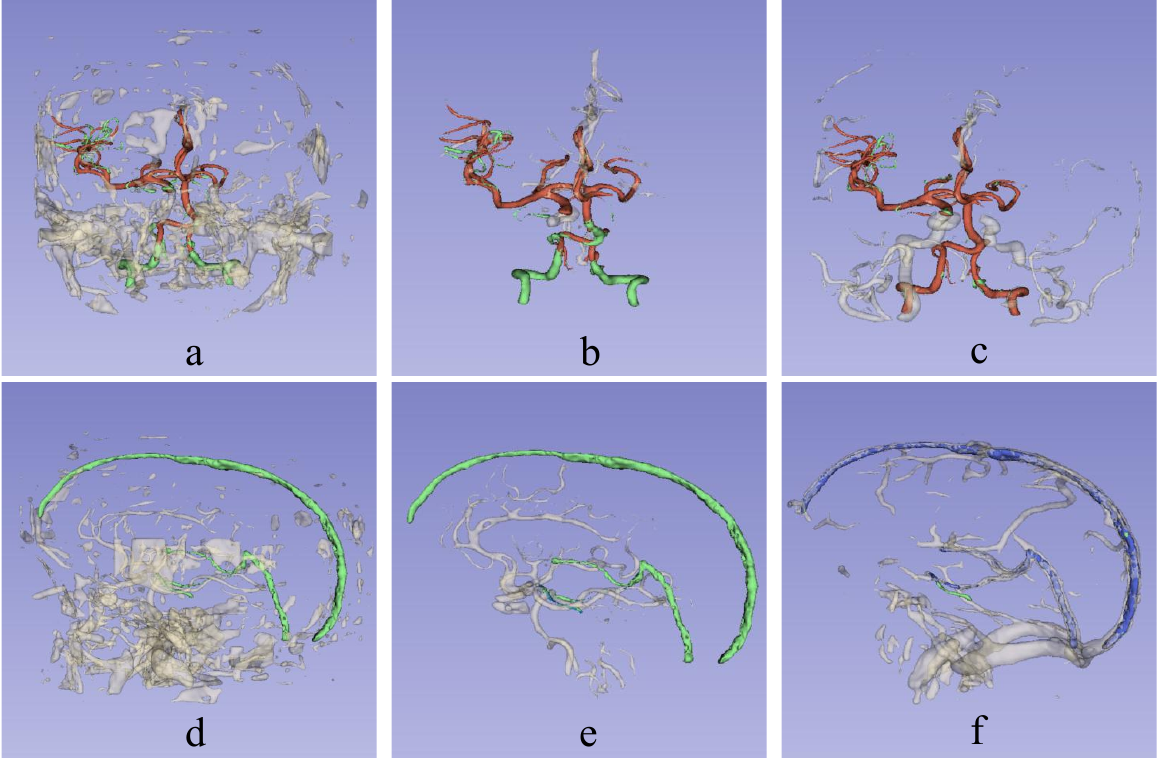}
   \caption{Qualitative comparisons of the segmentations from several models on a TopBrain sample. In the top row, we display results for artery segmentation using (a) VesselFM, (b) the VesselVerse nnUNet, and (c) our $M_{DV}$; true positive (TP) areas are \textbf{\textcolor{red}{red}}, false positives (FPs) are gray, and false negatives (FNs) are \textbf{\textcolor{ForestGreen}{green}}.  The bottom row includes the results for vein segmentation using (d) VesselFM, (e) the VesselVerse nnUNet, and (f) our $M_{DV}$; TP  areas are \textbf{\textcolor{blue}{blue}}, FPs are gray, and FNs are  \textbf{\textcolor{ForestGreen}{green}}. VesselFM and the VesselVerse nnUNet only output a single vessel class, so arterial predictions show up in the vein segmentation visualization as FPs, and vice-versa. We note that most FP areas segmented by $M_{DV}$ (c, f) are actually TP arterial and venous structures that are not annotated in TopBrain. Visualizations created on 3D Slicer. \label{fig:qual}     }  
    \end{center}
\end{figure}

We evaluate all four models on several test datasets, including held-out subsets of DynaVessel, TopCoW \cite{yangBenchmarkingCoWTopCoW2025}, and VesselVerse \cite{falcettaVesselVerseDatasetCollaborative2026a}. In addition, we use a subset of ISLES 2024 \cite{riedel2025isles24realworldlongitudinal} annotated by the organizers of the TopCoW challenge \cite{yangBenchmarkingCoWTopCoW2025}, and TopBrain, a subset of TopCoW with detailed annotations of 34 arteries and 6 venous structures. Since DynaVessel includes images acquired at multiple contrasts for each patient, we define two  evaluation variants: DynaVessel$_{1}$, which contains a single, arterial-phase image for each patient (14 images total), and DynaVessel$_{+}$, which contains all available contrast phases  per patient (165 images in total), allowing us to measure model robustness to contrast variability. Table \ref{tab:splits} offers more details about the datasets and the train/eval splits we used. Model performance is assessed using the metrics defined in Section \ref{sec:eval} in three main ways, namely: a) the binary vessel segmentation setting, where classes in the GTs and the predictions are merged into 0 = background, 1  = vessel; b) the artery vs vein segmentation setting, where classes in the GTs are merged into 0 = background, 1 = artery, 2 = vein; and c) the semantic segmentation setting, where we evaluate performance on all available arterial anatomical locations.

Importantly, for models that output a generic vessel segmentation, we use the entire vessel mask to separately measure artery and vein segmentation performance. Since our metrics prioritize sensitivity, this approach is biased toward models that output a single label, whereas models that output an artery vs vein mask, such as $M_{DV}$, will be penalized when they mistake arteries for veins and vice-versa. We follow a similar approach for semantic segmentation evaluation: we use the best available segmentation (artery, vein, or generic vessel) as the prediction to compute metrics when segmenting specific anatomical regions. 

In Table \ref{tab:results_1p}, we present the results for binary vessel segmentation. Despite being trained on a similarly-sized dataset (or much smaller, when compared to VesselFM), $M_{DV}$  achieves higher performance on the three datasets. Remarkably, training on DynaVessel leads to better results even on data that is \textit{in-domain} for the other models\footnote{The TopCoW, TopBrain, and VesselVerse test sets are in-domain for VesselFM and the nnUNet models trained on TopCoW/VesselVerse, and out-of-domain for $M_{DV}$. }. When the results are disaggregated to arteries and veins separately (see Table \ref{tab:artery_vein_results}), we see that $M_{DV}$ achieves higher performance on arteries and much higher performance on veins on TopBrain and DynaVessel, underlining the positive impact of our annotation protocol across the board. The metrics for $M_{DV}$ on artery segmentation are worse on DynaVessel$_+$ compared to the single-phase test set, which is expected as the former includes scans taken before contrast and at the early arterial, venous, and late venous phases. Fig. \ref{fig:detailed_phases} also reveals that the average scores are brought down by low performance on pre-contrast and early arterial phases, which is expected since at that point there is likely not enough contrast in the arteries. 

Fig. \ref{fig:semantic} shows a significant gap between the performance of $M_{DV}$ versus the two strongest baselines in the semantic segmentation evaluation, with great coverage of most anatomical regions annotated in TopBrain except for some distal arteries (the right and left anterior choroidal arteries). A similar trend is seen in the DynaVessel$_+$ test set, where the model generally performs worse on second-order arteries compared with the first-order ones. Meanwhile, in Fig. \ref{fig:qual}, we provide a qualitative visualization of the outputs produced by all three models. $M_{DV}$ achieves almost complete coverage of arterial and venous structures, with very few false negatives. The difference in performance is particularly noticeable on veins, where VesselFM and the VesselVerse nnUNet fail to detect most veins. In VesselVerse's case, this is due to the fact venous sinuses were not annotated to maintain consistency with the MRA annotations. Although a large number of false positive detections are visible in Fig. \ref{fig:qual}(b, c, e, f), the majority are correctly segmented arterial and venous structures that are not annotated in TopBrain. On the other hand, Fig. \ref{fig:qual}(a, d) shows that VesselFM generates a much larger number of spurious segmentations and unconnected elements, likely due to the limited amount of CTA it was trained on.

\subsection{Ablation study}

\begin{figure*}[h!]
        \subfloat{
            \includegraphics[width=.50\linewidth]{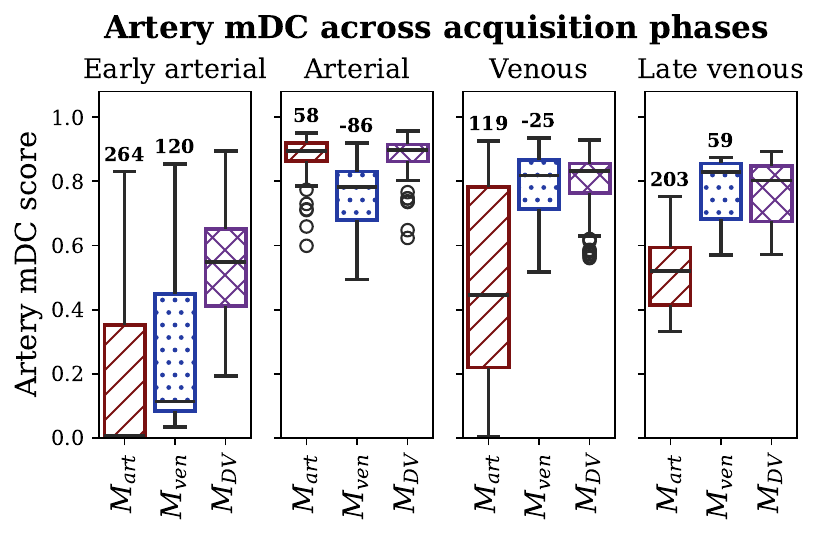}%
            \label{subfig:a}%
        }\hfill
        \subfloat{
            \includegraphics[width=.50\linewidth]{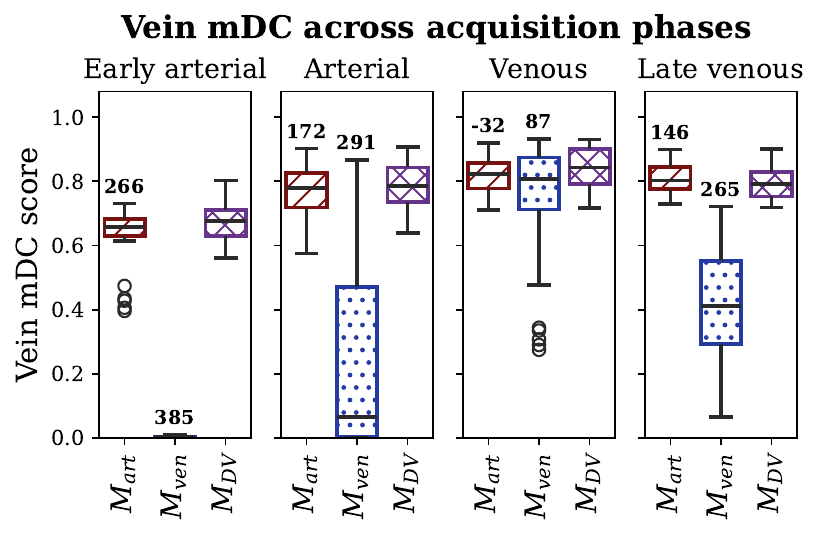}%
            \label{subfig:b}%
        }
        \caption{Artery and vein mDC scores across contrast phases for our model  ($M_{DV}$, \textcolor{violet}{\textbf{violet}}, rightmost box in each subplot), compared with models trained either on the peak arterial phase images ($M_{art}$, \textcolor{red}{\textbf{red}}, leftmost on each box) or the peak venous phase images ($M_{ven}$, \textcolor{blue}{\textbf{blue}}, middle box in each subplot). The number above the top whisker for each boxplot codifies the difference between the average Hounsfield unit (HU) intensity in arteries/veins during training vs during the phase on which the model is evaluated. The performance for single-phase training correlates negatively with the difference between mean training HU and mean test HU. The behavior in the late venous phase is explained by the circulation of part of the contrast agent back into arteries. } \label{fig:ablation}
    \end{figure*}

In Fig. \ref{fig:ablation}, we measure the impact of removing the various contrast phases used for training $M_{DV}$ by optimizing two new models: $M_{art}$, which is trained on only 29 peak arterial phase images, and $M_{ven}$, which is trained on only 29 peak venous phase images. It is important to note that $M_{art}$ and $M_{ven}$ have access to the same number of patients and GT annotations as $M_{DV}$. We make two observations: 1) that training on a single, peak phase per patient significantly reduces the models' capabilities to generalize across unseen phases; and 2) that the model trained on all available phases often surpasses the models trained on a single phase when evaluated on such phases. This once again confirms the usefulness of our methodology, especially in scenarios where the availability of peak arterial/venous phase images is not guaranteed.

\section{Discussion}

We proposed an automatic data pre-processing method that utilizes dynamic CTA images acquired at multiple time points. This approach not only reduces annotation time but also enhances the diversity of the training data at no additional cost. Using this protocol, we created DynaVessel, a CTA intracranial vessel segmentation dataset with artery and vein labels comprising 110 images from 29 patients, covering a wide range of acquisition contrasts. Furthermore, we annotated a large evaluation set of 165 images from 14 patients with fine-grained semantic labels for veins and 20 arteries, and we evaluate the impact of training $M_{DV}$ (a state-of-the-art nnUNet) on our data using modified metrics to emphasize sensitivity across different vessel segments. Overall, our evaluation pipeline provides detailed insight on the model’s ability to segment arteries of varying sizes and across multiple phases.

The $M_{DV}$ performance trends across dynamic CTA phases highlight the dependence of deep learning–based vessel segmentation on image contrast. In our experiments training on a single phase, segmentation performance degrades severely during mismatched phases—such as arterial segmentation in late venous images or venous segmentation in early arterial images—where reduced contrast conspicuity introduces ambiguity. Notably, the difference in HU distributions between the training data and phase-specific test data correlates strongly with  performance; for instance, $M_{ven}$ is trained on images with a mean vein HU of 485, resulting in a delta of 87 HU with respect to the mean vein HU of the venous phase ($mDC = 0.81$) and a delta of 385 HU with respect to the mean HU of the early arterial phase ($mDC = 0.01$). Together, these findings suggest that models trained on single-phase data leverage contrast-driven vessel appearance rather than anatomical priors. On the other hand, models trained on multi-phase data suffer far less performance degradation, although they are still sensitive to changes in contrast. Indeed, the artery segmentation performance of $M_{DV}$ peaks during the arterial phase, when arterial enhancement is maximal and venous structures remain relatively unopacified, whereas vein segmentation achieves its highest accuracy during the venous phase, when venous enhancement dominates. Moreover, $M_{DV}$ still requires the presence of some contrast agent to correctly segment brain vessels, as shown by its relatively poor performance during the early arterial phase. 
 

Regardless, when evaluated across all vascular regions, $M_{DV}$ achieved superior performance compared to models trained on other datasets of similar or much larger sizes. This performance gap was especially pronounced in veins, where $M_{DV}$ achieved a modified DC of 0.75 to 0.95 depending on the dataset, compared to under 0.01 mDC for the other models. The topology Sensitivity and average directed Hausdorff Distance tell a similar story, with $M_{DV}$ outperforming the other models by a significant margin. Although the performance gap was smaller for arteries, $M_{DV}$ still came on top, with modified DC scores of 0.75 to 0.86 compared to scores ranging from 0.07 to 0.45 for the other models. Overall, our work constitutes a significant step toward the automatic segmentation of all brain vascular structures in CTA scans, with broad applicability as a primary or auxiliary input for the assessment of cerebrovascular pathologies. Nevertheless, despite the fact DynaVessel imbues $M_{DV}$ with robust vessel segmentation performance, we note that our fine-grained semantic segmentation evaluations only serve to provide an upper-bound approximation of the performance of models trained using DynaVessel. Future work should address the semantic segmentation scenario by expanding the annotations in the training data to include all 20 artery (or more) sub-classes, establishing guidelines to annotate venous structures in a fine-grained manner, and designing deep learning architectures suited for such a complex task.

\begin{credits}
\subsubsection{\ackname} The authors acknowledge the financial support provided by NIH
grant 1R01LM013891-01A1. Ceballos-Arroyo, A. is grateful for the funding provided by Colombia’s Minciencias and Fulbright under the Fulbright Minciencias 2021 program.

\subsubsection{\discintname}
The authors have no competing interests to declare that are
relevant to the content of this article. 
\end{credits}

\newpage
\bibliographystyle{splncs04}

\newpage

\section*{Supplementary material}
\beginsupplement

\begin{figure}[ht]
   \begin{center}
   \includegraphics[width=\textwidth]{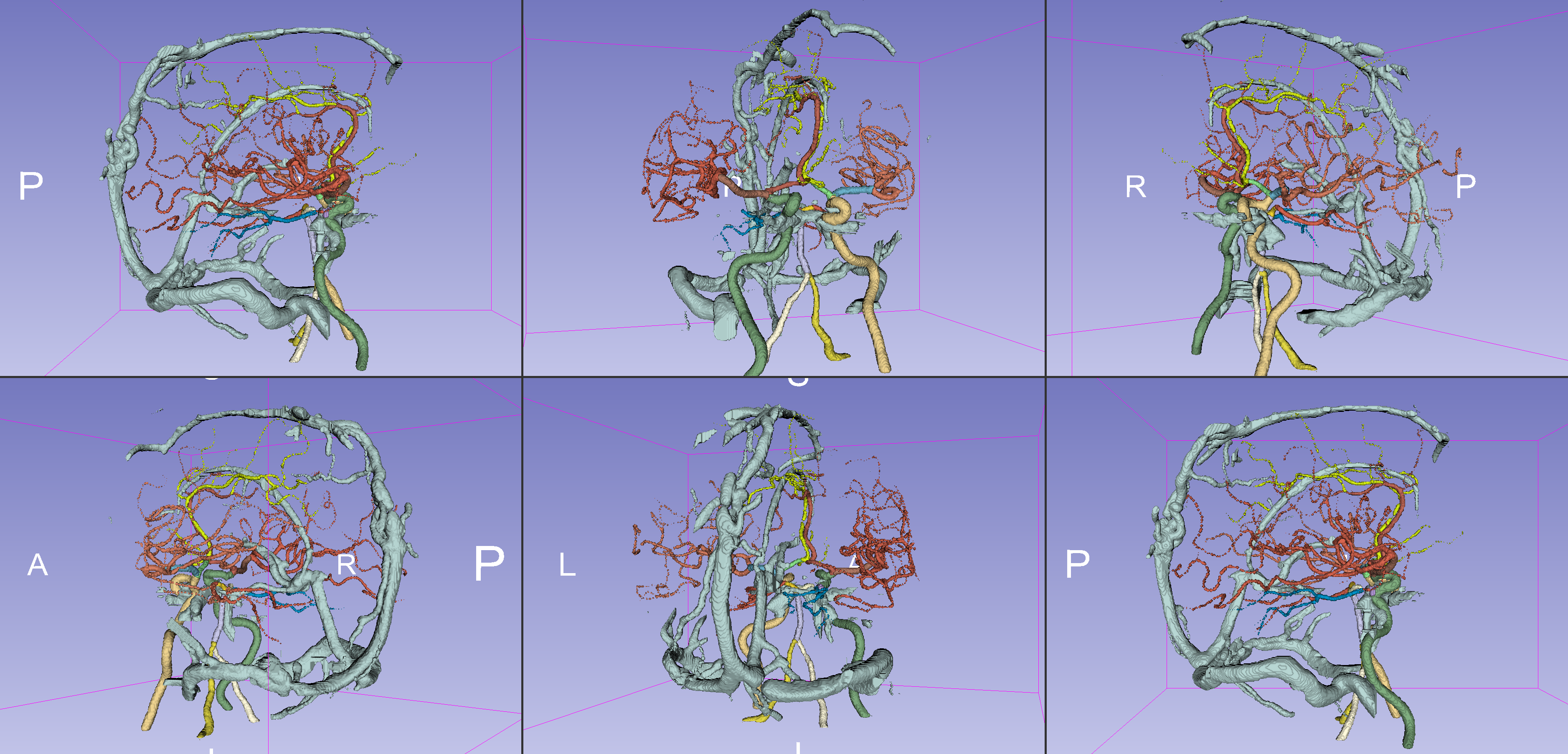}
   \caption{Visualization of the semantic labels for the DynaVessel test data. Created using 3D Slicer.
   \label{fig:gt} 
    }  
    \end{center}
\end{figure}

    \begin{figure*}[h]
        \subfloat[$X_{a}$]{%
            \includegraphics[width=.18\linewidth]{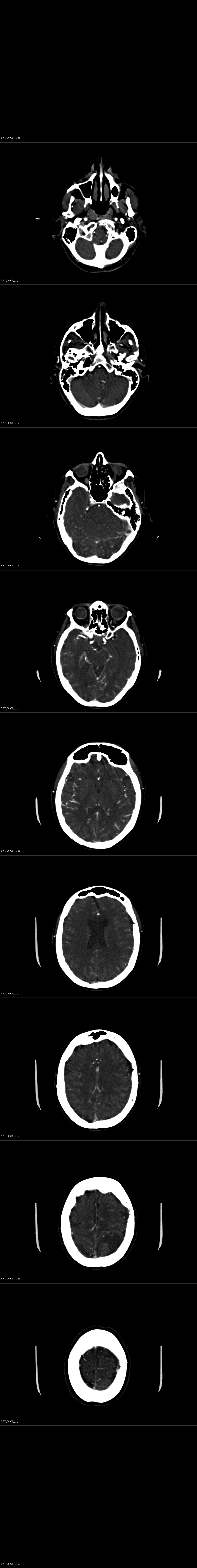}%
             \label{subfig:pc}%
        }\hfill
        \subfloat[$S_{a}$]{%
            \includegraphics[width=.18\linewidth]{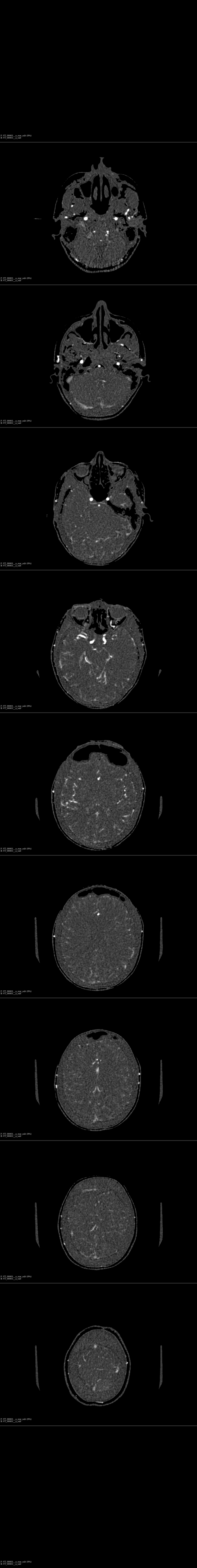}%
            \label{subfig:a_sub}%
        }\hfill
        \subfloat[$S_{a}^*$]{%
            \includegraphics[width=.18\linewidth]{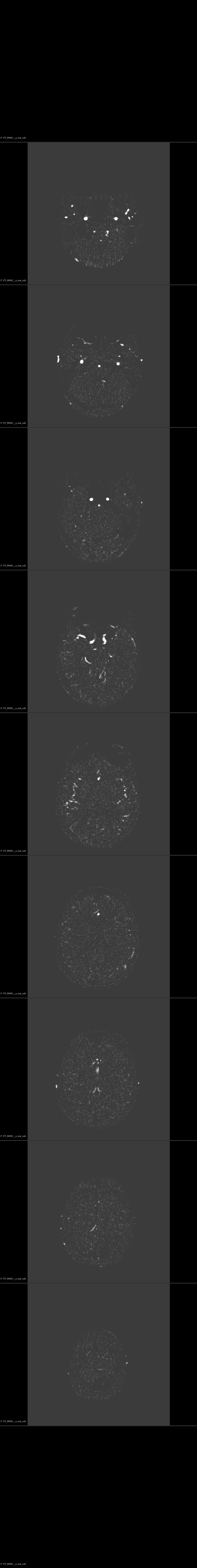}%
            \label{subfig:a_star_sub}%
        }\hfill
        \subfloat[$S_{v}$]{%
            \includegraphics[width=.18\linewidth]{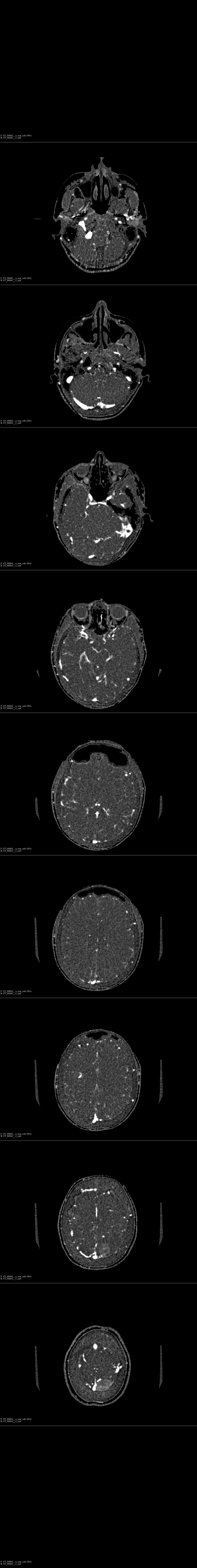}%
            \label{subfig:v_sub}%
        }\hfill
        \subfloat[$S_{v}^*$]{%
            \includegraphics[width=.18\linewidth]{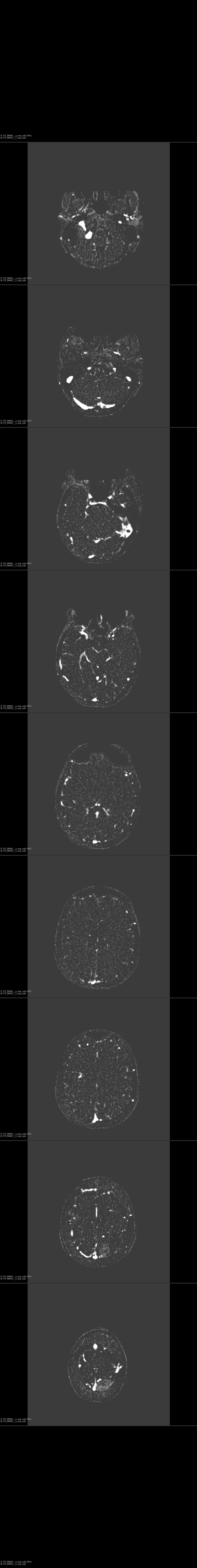}%
            \label{subfig:v_sub}%
        }\hfill
        \caption{Axial images of arterial phase CTA ($X_{a}$), bone subtracted arterial and venous phase ($S_{a}$ and $S_{v}$), and the vessel separated arterial and venous phase ($S_{a}^*$ and $S_{v}^*$) volumes from a sample patient.} \label{fig:phases_example}
    \end{figure*}

\end{document}